\documentclass{article}
\usepackage{float}
\usepackage{amsmath}
\usepackage{graphicx}
\usepackage[table]{xcolor}
\usepackage{makecell}

\PassOptionsToPackage{numbers, compress}{natbib}

\usepackage[dblblindworkshop, final]{neurips_2026}   
\workshoptitle{TAE (Trust-AI-Eval): Can We Trust AI Evaluation?}

\usepackage{multirow}
\usepackage[utf8]{inputenc} 
\usepackage[T1]{fontenc}    
\usepackage{hyperref}       
\usepackage{url}            
\usepackage{booktabs}       
\usepackage[most]{tcolorbox}
\usepackage{amsfonts}       
\usepackage{nicefrac}       
\definecolor{promptbg}{HTML}{F6F7FA}      
\definecolor{promptframe}{HTML}{C3CBD9}   

\definecolor{prompttitle}{HTML}{2B3A55}   
\definecolor{promptstrip}{HTML}{4A6188}   
\definecolor{promptaccent}{HTML}{1F3864}  
\definecolor{promptcode}{HTML}{ECEFF4}    
\definecolor{userbg}{HTML}{FBF7F0}        
\definecolor{userframe}{HTML}{DCC9A6}     
\definecolor{usertitle}{HTML}{6B5327}     
\usepackage{microtype}      
\usepackage[framemethod=default]{mdframed}

\newmdenv[
  linewidth=0.4pt,            
  linecolor=black,
  innerleftmargin=7pt,        
  innerrightmargin=7pt,
  innertopmargin=7pt,
  innerbottommargin=7pt,
  userdefinedwidth=0.93\linewidth,
  align=center,
  skipabove=0pt,
  skipbelow=0pt
]{promptbox}
\usepackage{framed}

\definecolor{promptbg}{HTML}{F5F7FA}      
\definecolor{promptframe}{HTML}{B9C3D4}   
\definecolor{prompttitle}{HTML}{2B3A55}   
\definecolor{promptaccent}{HTML}{1F3864}  
\definecolor{promptcode}{HTML}{E8ECF2}    
\definecolor{userbg}{HTML}{FBF6EE}        
\definecolor{userframe}{HTML}{D8C4A0}     
\definecolor{usertitle}{HTML}{6B5327}     

\newcommand{\boxheader}[2]{%
  \par\noindent
  {\setlength{\fboxsep}{4pt}%
   \colorbox{#1}{\parbox{\dimexpr\linewidth-8pt}{%
     \color{white}\footnotesize\sffamily\bfseries\raggedright #2}}}%
  \par\medskip}

\usepackage{xcolor}         
\usepackage{wrapfig}
 
\usepackage{array}
\usepackage{subcaption}
\usepackage{longtable}
\newcommand{\follow}[1]{\textcolor{blue!75!black}{\textbf{#1}}}
\newcommand{\pass}[1]{\cellcolor{green!30}\textbf{#1}}

\title{It's Not What the Image Shows: Irrelevant Context Destabilises VLM Judges Without Informing Them}

\author{%
  \textbf{Nagham Omar}\thanks{Equal contribution.} \quad
  \textbf{Mahmoud Jabarin}\footnotemark[1] \quad
  \textbf{Kinan Ibraheem}\footnotemark[1] \quad
  \textbf{Lotem Peled-Cohen} \\[2pt]
  Faculty of Data and Decision Sciences, Technion \\[2pt]
  \texttt{\{nagham.omar, mahmoud.j, kinani, splotem\}@campus.technion.ac.il}
}

\hypersetup{
  colorlinks=true,
  linkcolor=black,
  citecolor=black,
  urlcolor=blue,
  filecolor=blue
}

\begin{document}

\maketitle

\begin{abstract}
Vision-language models (VLMs) are increasingly used in place of human
annotators, making it important that substitutability tests reflect the model
rather than incidental evaluation conditions. We introduce
\textsc{MIST}, the Misleading-Image Stress Test: 200 English
sentences, each built around a phrase readable either figuratively or literally
and shown with an \emph{aligned} image depicting its reading, a
\emph{misleading} image depicting the opposite, or no image at all. The
guidelines require the label to be decided from the sentence alone, so no image
should change any answer. We expected each image to pull a judge's labels toward
the sense it depicts, and neither kind did. Across thirteen VLM judges, an
aligned image changed 20.5\% of labels and a misleading one 19.4\%, close for
every judge and both above the 11.6\% produced by deleting the
ignore-the-image instruction with the image left in place. Yet only 37\% of the
labels that differ between the two images moved toward the sense shown, and
agreement with our human annotators is unchanged whether the image is absent,
aligned or misleading. The effect is smaller in the seven judges that pass the
alt-test than in the six that never do, but present in all of them: what moves a
judge is that an image is there, not which of the two it is, so a
substitutability verdict describes a configuration as much as a model.
\end{abstract}

\section{Introduction} \label{sec:intro}

Large language models (LLMs) increasingly annotate in place of humans
\citep{gilardi2023chatgpt, zheng2023judging}, and several protocols decide when
that substitution is justified. The alternative annotator test
\citep[alt-test;][]{calderon2025alternative} asks whether an LLM judge agrees
with a panel of human annotators at least as well as a withheld member does;
\citet{he2025hide} ask whether its labels are statistically indistinguishable
from a human's. Each returns one verdict per judge from a single run: one
prompt, one format, one presentation of the item. But real items arrive wrapped
in context, an image or a preceding message, that the guidelines declare
irrelevant. Human annotators can usually be instructed to ignore such
information; models might not. Existing substitutability verdicts do not reveal
whether their conclusions are robust to these changes in context, so we ask
whether a verdict describes the judge or the conditions it was measured under.

The stakes are high: once a human panel is replaced its labels become evaluation
sets and training data, and errors in them propagate into flawed comparisons and false conclusions
\citep{nahum2025llms}. Judges are already known to favour longer answers and
their own outputs \citep{zheng2023judging}, shift with the scoring format
\citep{chen2024mllm} and change with the prompt \citep{li2025generation}, but
this is treated as an accuracy problem to engineer away rather than a threat to
the decision that a model may replace a person.

Testing it needs a task where the context can change while the correct answer
stays put, and idioms provide one. An idiom is a multiword expression whose
meaning is not composed from its parts \citep{villavicencio2005introduction}: to
\emph{kick the bucket} is to die, and no kicking or bucket is involved. Being an
idiom is a property of the string; how it is read is a property of the sentence,
and many idioms keep a usable literal one, so the same string is
\emph{figurative} in \emph{my grandfather kicked the bucket last winter} and
\emph{literal} in \emph{she kicked the bucket over and spilled the water}. An
image can therefore be placed beside such a sentence, and swapped, without
touching the answer.

Existing multimodal work is built the opposite way: IRFL \citep{yosef2023irfl}
and AdMIRe \citep{pickard2025semeval} make the image the object of the decision,
so changing it legitimately changes the answer. We instead distinguish models
that use context appropriately from those whose judgments are influenced by
context they should ignore. ID10M-JAM \citep{hashiloni2026id10m} also preserves
the label, but perturbs text rather than image and scores identification
accuracy. Work on irrelevant context is closer \citep{shi2023large,
gonen2025semantic, deng2025words} but measures task accuracy, not agreement with
the human annotators a model would replace.

We introduce \textbf{\textsc{MIST}}, the Misleading-Image Stress Test: 200
English \emph{items}, each a sentence with one potentially idiomatic phrase, the
\emph{target}, whose \emph{label} records how that target reads in that
sentence. The guidelines require the label to be decided from the sentence
alone, so a good annotator, human or VLM, gives an item
the same label whether the image beside it depicts the target's figurative
reading, its literal one, or is absent. Swapping the image is therefore
label-preserving by construction\citep{ribeiro2020beyond}: the human annotator or VLM judge is perturbed,
the correct answer is not, and any change of label is an error. Both images
depict a reading of the target, so what varies is which reading is shown, not
whether the image is about the sentence at all.
Our results establish two points. First, adding an image changes a judge's
labels even when the prompt explicitly instructs it to disregard the image.
Second, the effect of image content is weaker than expected: aligned and
misleading images move similar numbers of labels, and neither reliably shifts
labels toward the interpretation it depicts. We release \textbf{\textsc{MIST}}
with all human annotator and VLM judge labels,\footnote{\url{https://huggingface.co/datasets/naghamo/mist-vlm-judges}} and the finding that context a VLM
judge is told to ignore destabilises it without informing it. Related work is in
Appendix~\ref{sec:related}.

\section{\textsc{MIST}: The Misleading-Image Stress Test} \label{sec:mist}

\begin{figure}[t]
\centering
\captionsetup[subfigure]{labelformat=empty}
\begin{subfigure}[t]{0.24\linewidth}
  \centering
  \includegraphics[width=0.85\linewidth]{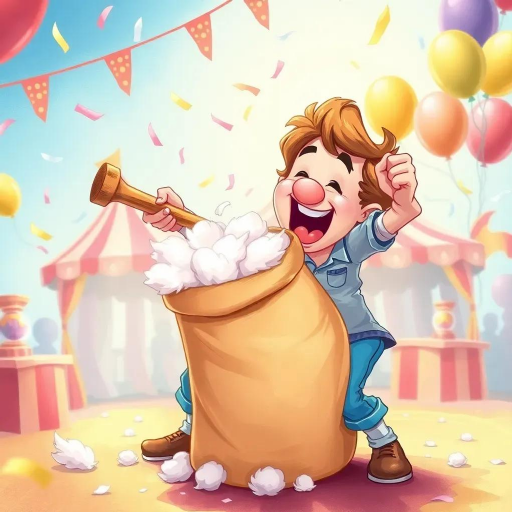}
  \caption{ \textcolor{red!70!black}{\textbf{T1}}: figurative sentence,
  literal image. \textit{Exhausted, John was ready to \textbf{hit the sack}.}}
\end{subfigure}\hfill
\begin{subfigure}[t]{0.24\linewidth}
  \centering
  \includegraphics[width=0.85\linewidth]{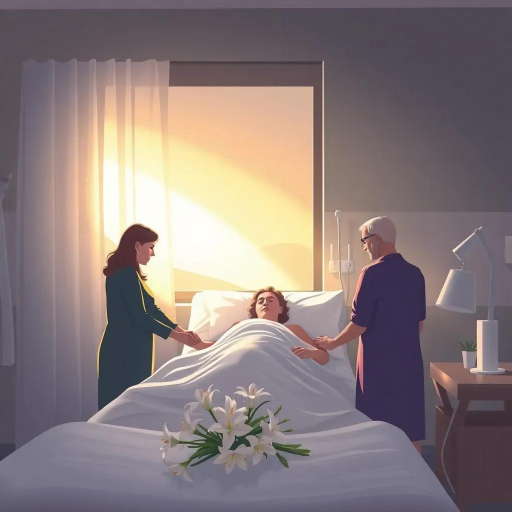}
  \caption{ \textcolor{green!50!black}{\textbf{T2}}: figurative sentence,
  figurative image. \textit{After a long illness, he \textbf{kicked the
  bucket}.}}
\end{subfigure}\hfill
\begin{subfigure}[t]{0.24\linewidth}
  \centering
  \includegraphics[width=0.85\linewidth]{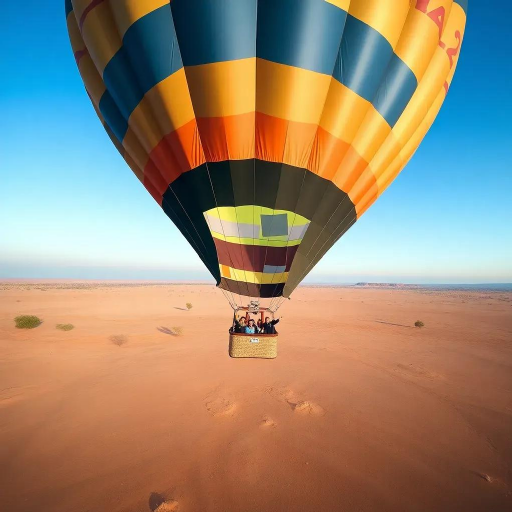}
  \caption{ \textcolor{green!50!black}{\textbf{T3}}: literal sentence,
  literal image. \textit{After the balloon ride, we were back \textbf{down to
  earth}.}}
\end{subfigure}\hfill
\begin{subfigure}[t]{0.24\linewidth}
  \centering
  \includegraphics[width=0.85\linewidth]{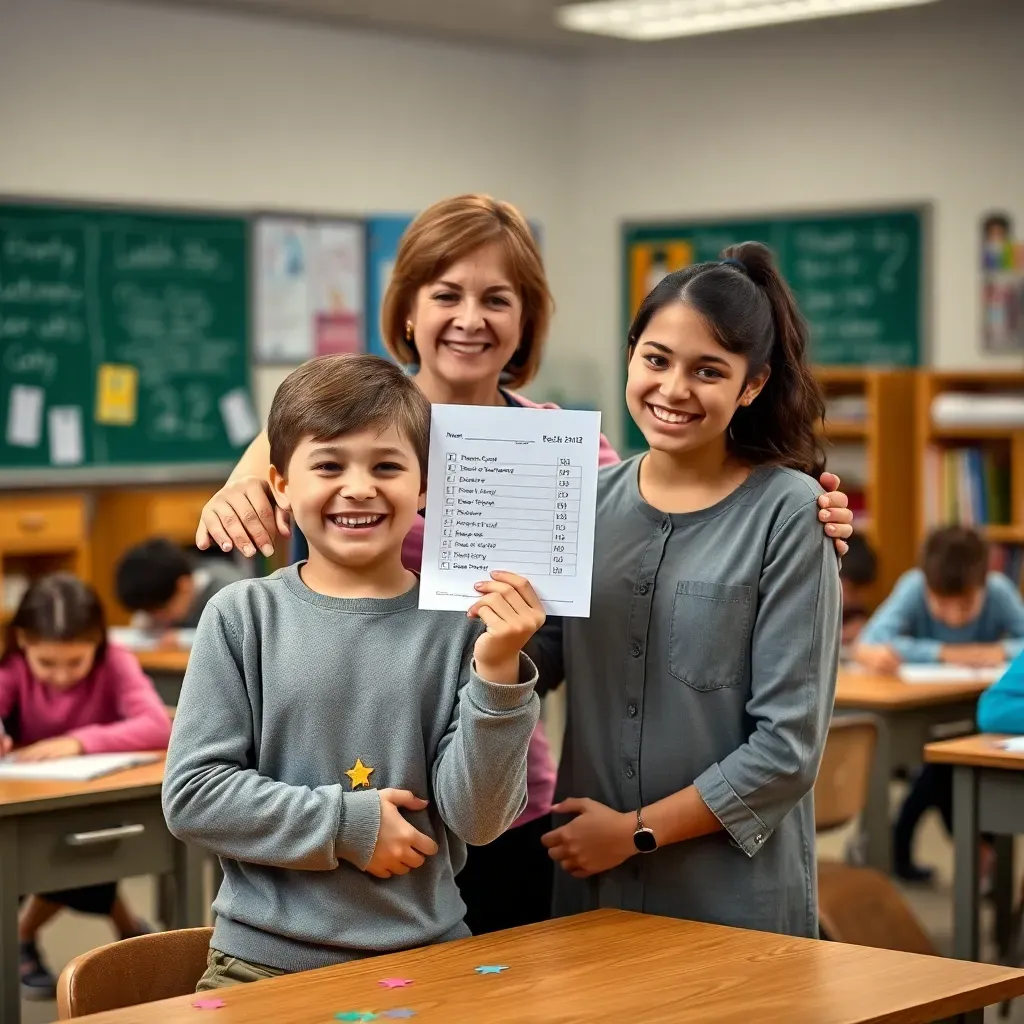}
  \caption{ \textcolor{red!70!black}{\textbf{T4}}: literal sentence,
  figurative image. \textit{The \textbf{teacher's pet} parakeet sat on her
  shoulder as she graded.}}
\end{subfigure}
\caption{The four conditions of \textsc{MIST}, 50 phrases each. T2 and T3 are
\emph{aligned}, T1 and T4 \emph{misleading}. \textbf{Bold} marks the target.}
\label{fig:conditions}
\end{figure}

\paragraph{Source and construction.} We build on a public instruction-tuning
release derived from the AdMIRe shared task \citep{pickard2025semeval}, holding
551 English potentially idiomatic expressions. Each expression appears in two
rows: one sentence using it figuratively and one using it literally, each paired
with a generated image of that reading. We sample 200 expressions and keep one
sentence from each, 100 figurative and 100 literal; we call the reading that
sentence uses its \emph{sentence sense}. Because the discarded row's image
remains available, every retained sentence can be shown with either image
(Appendix~\ref{app:construction}).

\paragraph{Conditions.} An \emph{item} is a sentence together with its target
expression and its sentence sense. Each item is presented in three inputs: with
an \emph{aligned} image, which depicts the sentence sense; with a
\emph{misleading} image, which depicts the opposite one; or with no image.
Crossing sentence sense with image type yields the four conditions of
Figure~\ref{fig:conditions}, fifty expressions each. Each human annotator saw
one condition per expression, so the manipulation could not be inferred by
comparing versions; the VLM judges saw all three inputs of every item.

\paragraph{Labels.} The label describes how the written phrase reads in its
sentence, not what the image shows. Two questions assign one of four labels, and
they ask about different things. The first is about this sentence: does the
phrase carry its literal meaning here? If it does and no figurative reading is
present, the label is \emph{Fully Literal} (LL); if a figurative reading is also
present, \emph{Figurative and Literal} (FL). If it does not, the second question
sets the sentence aside and asks about the phrase itself: does any semantic link
remain between its literal words and its figurative meaning? If so the label is
\emph{Weak Figurative} (WF), otherwise \emph{Fully Figurative} (FF). We treat
the labels as partially ordinal, from FF to LL, most to least figurative. The
guidelines add that a literal image does not by itself make a phrase Fully
Literal, and tell anyone distracted by the image to annotate as if it were
hidden (Appendix~\ref{app:guidelines}).

\paragraph{Human annotation.} Six fluent English speakers annotated
\textbf{\textsc{MIST}} as two disjoint trios. One knew the study design and was
experienced with the task (\emph{informed}), the other received only the
guidelines (\emph{blind}). Within each trio, three annotators independently
labelled the same 100 items, with no item labelled by both trios. Both reach
moderate agreement and do not differ significantly (Fleiss $\kappa$ 0.65 and
0.57, $p = 0.22$): the task is hard in itself, not because either trio was
misled.

\section{Experiments} \label{sec:exp}

\subsection{Experiments Setup} \label{sec:setup}

\paragraph{VLM Judges.} We evaluate thirteen VLMs as candidate annotators, four
proprietary and nine open-weight, spanning $3$B to $32$B active parameters.
Seven pass the alt-test in at least one configuration and are the ones the body
reports, covering five families: GPT-5.2 \citep{openai2025gpt52},
Gemini~3.1~Flash-Lite (G3.1-FL) and Gemini~3.5~Flash (G3.5-F)
\citep{google2025gemini3}, Gemma-3-27B (Gm3-27B) \citep{gemmateam2025gemma3},
Mistral-Small-3.2-24B (MiS-24B) \citep{mistral2025small32}, and Qwen3.6-27B
(Q3.6-27B) and Qwen3.6-35B-A3B (Q3.6-35B) \citep{qwen2026qwen36}. The other six
never pass and are named, cited and reported in
Appendix~\ref{app:change-full}. Where a model offers a reasoning mode we disable
it, keeping explicit reasoning a controlled prompt factor.

\paragraph{Prompts and image instructions.} VLM judges are sensitive to
prompting \citep{li2025generation}, so each runs under four techniques sharing
one task instruction and one output contract: \emph{zero-shot},
\emph{few-shot}, \emph{chain-of-thought} (CoT) \citep{wei2022chain} and
\emph{few-shot with CoT} (Appendix~\ref{app:prompt}). We cross each prompt with
an \emph{explicit} arm that tells the judge to ignore the image and a
\emph{silent} arm that never mentions it. Each judge produces
$4 \times (1 + 2 \times 2) = 20$ labels per item, 52{,}000 in total, decoded
greedily with the output constrained to the four labels. Unless stated otherwise
we report the explicit arm, pooled over the four prompts; the silent arm shows
the same pattern with different labels (Table~\ref{tab:change-main}, column~3).

\paragraph{Measures.} We report four quantities. \textbf{(i)} A \emph{change rate} is the
proportion of (item, prompt, judge) cells whose label differs between two
inputs: from text-only to the same item with an aligned image and, separately,
with a misleading one. Then as a \emph{control}, we hold the image fixed and
delete only the paragraph instructing the judge to ignore the image, giving an
upper bound on how much a prompt edit alone can move a judge. \textbf{(ii)}
\emph{Direction} is defined on the cells whose label differs between the two
images: the proportion moving toward the sense shown, where chance is 50\%.
\textbf{(iii)} \emph{Agreement} is exact match with the human majority label of the trio
that saw the item. \textbf{(iv)} \emph{Substitutability} is the alt-test, run per trio
because the trios share no items, at the slack prescribed for each tier
($\varepsilon = 0.20$ informed, $0.15$ blind); a judge is scored only on the
input its human counterparts saw.

\subsection{Results} \label{sec:results}

\begin{table}[t]
\centering
\caption{Change rates, pooled over the four prompts, for the seven judges that
pass the alt-test in at least one configuration.}
\label{tab:change-main}
\footnotesize
\setlength{\tabcolsep}{5pt}
\begin{tabular}{@{}lrrrr@{}}
\toprule
& \multicolumn{3}{c}{\textbf{labels that change (\%)}} & \\
\cmidrule(lr){2-4}
\textbf{Judge}
& \makecell[r]{attaching an\\aligned image}
& \makecell[r]{attaching a\\misleading image}
& \makecell[r]{prompt edited,\\image fixed}
& \makecell[r]{of those that differ,\\\% toward the image} \\
\midrule
GPT-5.2    & 10.0 &  9.1 &  7.7 & 28 \\
G3.5-F     & 10.9 & 11.8 &  7.9 & 34 \\
G3.1-FL    & 14.5 & 13.0 &  8.4 & 12 \\
\midrule
Gm3-27B    & 17.9 & 19.2 & 10.6 & 30 \\
MiS-24B    & 21.2 & 16.9 &  9.1 & 21 \\
Q3.6-35B   & 16.6 & 14.2 &  9.6 & 39 \\
Q3.6-27B   & 20.0 & 18.5 &  7.9 & 38 \\
\midrule
\textbf{mean} & \textbf{15.9} & \textbf{14.7} & \textbf{8.8} & \textbf{29} \\
\bottomrule
\end{tabular}
\end{table}

\paragraph{Some judges are close enough to a human trio to be worth perturbing.}
Against the informed trio no judge passes the alt-test in any of the 52
judge-by-prompt cells; that trio agrees more closely with one another, so the
bar a withheld member sets is higher. Against the blind trio, two judges pass at
the slack prescribed for its tier, G3.5-F and Gm3-27B, and seven at the more
permissive $\varepsilon = 0.20$, while six never pass
(Appendix~\ref{app:alttest}). The seven the body reports are therefore those
that clear the test somewhere in the grid, not all of them at the prescribed
slack; they are the group for which it is meaningful to ask what perturbs them,
and Appendix~\ref{app:change-full} gives all thirteen.

\paragraph{Adding an image changes the labels a judge produces.} Attaching an
image to a sentence a judge has already labelled changes 15.9\% of its labels if
the image is aligned and 14.7\% if it is misleading
(Table~\ref{tab:change-main}); across all thirteen judges, 20.5\% and 19.4\%
(Appendix~\ref{app:change-full}). The two are close for every judge, never more
than five points apart, so what moves the label is that an image is present, not
which one it is. Both exceed the control, holding the image fixed and deleting
the ignore-it paragraph, for every judge individually and not merely on average:
telling a judge in plain language to disregard the image moves fewer labels than
placing the image there does.

\paragraph{The labels that change do not follow the image.} Across all thirteen
judges the two images disagree on 17.1\% of (item, prompt, judge) cells,
1{,}776 of 10{,}400. Of these, 37\% move toward the sense the misleading image
depicts and 63\% move away from it, and the imbalance holds for each sentence
type separately (683 figurative, 38\% following the image; 1{,}093 literal,
35\%). Every judge that passes the alt-test falls below chance, the highest at
39\%; across all thirteen, only one exceeds it, by two points. Nor does the
movement cross the distinction the task turns on: among the seven judges, 71\%
of it stays on the same side of the figurative--literal divide, so the image
reshuffles a judge's answer without changing which reading it believes.

\paragraph{The image does not reduce accuracy.} Agreement with the human
majority is 54.4\% with no image, 53.7\% aligned and 54.7\% misleading, and only
5 of 13 judges lose accuracy under an image. The image changes \emph{which}
items a judge agrees with the humans on, not \emph{how many}: each change is an
error by construction, but they cancel in the aggregate, so no measure computed
from overall agreement can see them. Agreement is lower on misleading items than
aligned ones, but that gap is largest with no image at all, so it reflects the
difficulty of the two disjoint expression sets rather than anything the image
did (Appendix~\ref{app:stats}).

\section{Discussion} \label{sec:discussion}

We expected each image to pull a VLM judge toward the sense it depicts, and neither does: attaching a picture moves 15.9\% of labels, the movement does
not track what the picture shows, and agreement with our human annotators is
unchanged. Misleading content moves a judge no more than agreeing content does,
which separates \emph{using} an image from \emph{being influenced by what it
depicts}. The effect is smaller in the seven judges that pass the alt-test than
in the six that never do, 15.9\% against 25.8\%, but each moves more under an
image than under a prompt edit, so the instability sits in the models a
practitioner would deploy.
A substitutability protocol cannot see this: aggregate agreement conceals
instance-level instability, so calling a judge substitutable describes a judge
and a configuration at once and names only the first. A deployment report should
state the prompt used and any irrelevant context attached.
\textbf{\textsc{MIST}} is an invariance test \citep{ribeiro2020beyond}, run
against a judge rather than a task model, perturbed in a second modality that
leaves the sentence untouched, and in two directions rather than one, which is
what lets us ask whether content or mere presence moves the label. It is
presence. Such a test reports one property, and a judge could pass it without
reading its input; ours do read it, agreeing with the human majority at 54.4\%.
Whether they also move when the sentence genuinely changes reading is the
complementary measurement, which \textsc{MIST} does not make
(Appendix~\ref{app:limitations}). Whether the effect is specific to the alt-test
and figurative annotation is the next question.

\bibliographystyle{plainnat}
\bibliography{ref}


\appendix

\section{Related Work}
\label{sec:related}

\paragraph{Judges and their measurement conditions.} LLM judges are swayed by
position, verbosity, and self-enhancement \citep{zheng2023judging}, and
multimodal judges diverge further from human preference under scoring and
ranking \citep{chen2024mllm}. Their scores are also known to shift with the
prompt, treated there as a design choice to be optimised rather than as a threat
to a substitutability decision \citep{li2025generation}. Whether a judge may replace a
panel outright has been formalized twice: by the alt-test, which asks if it
matches the panel at least as well as a withheld human
\citep{calderon2025alternative}, and by \citet{he2025hide}, who test statistical
indistinguishability. Both report that verdict for one input and one
configuration, and neither asks whether it holds under another. The test we run
is behavioural testing's invariance expectation, where a label-preserving
perturbation must leave the prediction unchanged \citep{ribeiro2020beyond}.
It is paired there with a directional one, in which a perturbation is expected
to move the prediction in a specified direction, and both are applied to task
models rather than to evaluators. \textbf{\textsc{MIST}} runs the first against a
judge, under a perturbation in a second modality, and reads it against a
substitutability verdict rather than a consistency score.

\paragraph{Unwanted context.} Where robustness to unwanted context has been
studied, it has been scored on task accuracy rather than on agreement with
humans: irrelevant text degrades reasoning \citep{shi2023large}, irrelevant
prompt content leaks into generation \citep{gonen2025semantic}, and VLMs
privilege text over image when the two conflict, in their case with the text
corrupted rather than the image \citep{deng2025words}. Closest to us, ID10M-JAM
\citep{hashiloni2026id10m} prefixes conflicting cues to potentially idiomatic
expressions that humans still read unambiguously, but its perturbation also
lengthens the input, its human baseline rests on two annotators and raw
agreement, and it scores identification accuracy rather than agreement with the
annotators a model would replace.

\paragraph{Figurative benchmarks.} MAGPIE \citep{haagsma2020magpie}, FLUTE
\citep{chakrabarty2022flute}, and DICE \citep{mi2025rolling} are text-only,
while IRFL \citep{yosef2023irfl}, V-FLUTE \citep{saakyan2025understanding}, and
AdMIRe \citep{pickard2025semeval} add images and include mismatched pairings.
In all of them the image is the object of the decision: which image fits, or
whether it entails the claim. In \textbf{\textsc{MIST}} the image is never judged, the
decision is about the text, and the guideline says to disregard it, which is
what lets us perturb a judge without moving the label it is scored against.

To our knowledge no prior work asks which of a judge's measurement conditions
its substitutability verdict actually depends on.

\section{Limitations}
\label{app:limitations}

\paragraph{Only two images, both phrase-relevant.} Both images depict a reading
of the target phrase, so we contrast two relevant images rather than a relevant
image against an irrelevant one. The aligned arm is the partial counterpart: an
image agreeing with the sentence should confirm the label a judge has already
given, and instead it moves as many labels as the contradicting one
(Table~\ref{tab:change-main}). What that leaves untested is whether an image
about nothing in the sentence would move as many again.

\paragraph{Partially ordinal scale.} We order the labels FF to LL, but the two
boundaries are different judgments: FF versus WF asks about the phrase type, FL
versus LL about the instance, so a one-step move is not the same quantity
everywhere. The net shifts in
Table~\ref{tab:drift-appendix} average across that difference. The change rates
do not, being exact-match, and the 71\% of movement that stays on one side of
the first-step boundary uses only the distinction the scale encodes cleanly.

\paragraph{No human change rate.} Each annotator saw one condition per
expression, by design, so the manipulation could not be inferred by comparing
versions; the cost is that no human labelled the same item twice and we cannot
say how often a person's label moves for reasons unrelated to the image. The
prompt-edit control supplies the within-judge version of that baseline
(\S\ref{sec:setup}), and every judge exceeds it, but the human figure remains
an assumption of this design rather than a measurement.

\paragraph{Scope.} All items are English, drawn from one corpus of potentially
idiomatic expressions, and every judge runs with reasoning modes disabled so that
explicit reasoning stays a controlled prompt factor. The effect nonetheless
appears in all thirteen judges across seven families and a tenfold parameter
range, so it is not a property of one model or one vendor.

\section{Annotation Guidelines}
\label{app:guidelines}

Human annotators and VLM judges received the same task description and the same
label definitions. The text below reproduces the guidelines.

\paragraph{Task.} Read a sentence containing a highlighted phrase and decide how
that phrase is used in the exact sentence given. An image is shown alongside the
sentence as silent context. The phrase is the annotation target, not the image.
Annotators are told not to use the image to change their interpretation.

\paragraph{Labels.} \emph{Fully Figurative} (FF): the phrase is completely
figurative and its literal meaning plays no role, as in \emph{the elderly man
finally \textbf{kicked the bucket}}. \emph{Weak Figurative} (WF): figurative
here, but a semantic or conceptual link survives between the figurative meaning
and the literal words, as in \emph{\textbf{racing against the clock}}, where a
clock measures time and the idiom evokes time pressure. \emph{Figurative and
Literal} (FL): both readings are genuinely active in this sentence, as in
\emph{John found himself with his \textbf{back against the wall}}, steadying
himself in a crowded bar. \emph{Fully Literal} (LL): the word-by-word sense
only, with no figurative reading activated, as in \emph{the chef placed the
cucumbers in a \textbf{pickle jar}}.

\paragraph{Decision procedure.} Annotators apply two steps in order.
\emph{Step 1}, which depends on the sentence: does the phrase carry its literal
meaning in this exact sentence? If yes and no figurative reading is present, the
label is LL; if yes and a figurative reading is also present, FL; if no,
continue. \emph{Step 2}, which sets the sentence and the image aside: does the
figurative meaning share any semantic or conceptual connection with the literal
words? If so the label is WF, otherwise FF. Step 2 is a judgement about the
phrase type rather than the instance, so FF and WF differ in the relation
between the two meanings, while FL and LL differ in whether the literal reading
is active here. The Step 1 boundary therefore separates $\{$FF,~WF$\}$ from
$\{$FL,~LL$\}$, which is the largest distinction the ordinal scale encodes.

\paragraph{Additional rules.} Annotations are grounded in the sentence exactly as
written, with no added or removed articles, no near-synonym substitution, and no
assumed change of grammatical form; a phrase that would be literal only under a
slight grammar change is not literal as written. Annotators read the full
sentence before deciding, since a single word elsewhere can make the literal
reading plausible or impossible. On the image, the guidelines state that a
literal image does not by itself make a phrase Fully Literal and a figurative
image does not change a label the sentence does not support, and instruct
annotators who find the image distracting to annotate as if it were hidden and
then check that the answer holds. Where two labels remain plausible, annotators
choose the one matching their primary interpretation; where the uncertainty
itself comes from the phrase genuinely supporting two readings, that is a signal
to choose FL.

\section{Construction Details}
\label{app:construction}

\paragraph{Compound sampling.} From the 551 English compounds of the source
release\footnote{\texttt{UCSC-Admire/idiom-SFT-dataset-561-2024-12-06\_00-40-30} on the
Hugging Face Hub.}, we first restrict to compounds that have both figurative and literal
rows in which the compound appears either as an exact substring or as a simple
morphological variant (up to three filler tokens between content words, for
example \emph{lose his marbles} for \emph{lose your marbles}); this is the
necessary condition for cross-pairing. From that eligible pool we draw 200
compounds uniformly at random with \texttt{random.seed(123)} and partition them
sequentially into four disjoint groups of 50, so each compound appears in
exactly one condition.

\paragraph{Sentence selection.} The sentence is drawn from a figurative row for
T1 and T2 and from a literal row for T3 and T4. Among the candidate rows we
prefer shorter sentences through a length-priority cascade, taking a sentence of
at most 20 words where available, then at most 25, then at most 30, then any
length as a fallback, and picking uniformly from the three shortest sentences in
the winning tier. Of the 200 final sentences, 164 (82\%) contain the compound as
an exact substring and 36 (18\%) as a morphological variant; none is missing the
compound.

\paragraph{Image pairing.} Each item takes the \texttt{correct\_image} of the
relevant row: the row whose sense the sentence uses for T2 and T3, and the
opposite row for T1 and T4. No further curation is performed, since that field
already marks the image depicting the corresponding sense. The images are
model-generated in the original release, across roughly fifty art styles
recorded in its \texttt{style} field; we generate none ourselves. The remaining
four images supplied per row are unused.

\begin{figure}[h]
\centering
\includegraphics[width=0.85\linewidth]{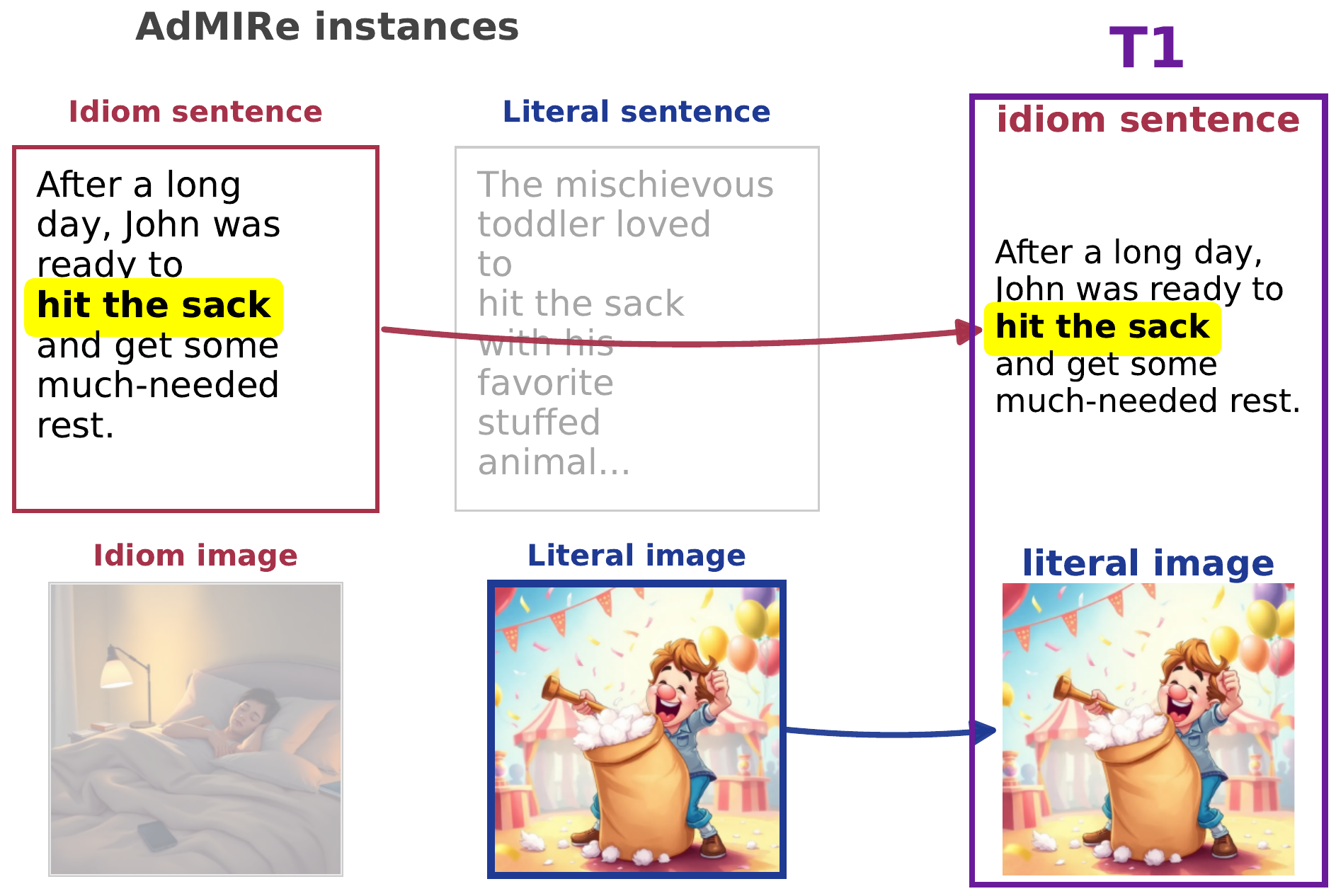}
\caption{Constructing a T1 item: the sentence comes from the compound's
figurative row and the image from the literal row's \texttt{correct\_image}. T2
and T3 take both from the same row; T4 mirrors T1 in the opposite direction.}
\label{fig:construction}
\end{figure}

\paragraph{Compound markup.} The compound is wrapped in \texttt{**\ldots**}
within the sentence, so that annotators and judges see the same marker on the
focal phrase. Marking uses the same exact-substring-then-variant cascade as the
eligibility filter; all 200 sentences were marked successfully.

\section{Data Statistics}
\label{app:stats}

\paragraph{Composition.} \textbf{\textsc{MIST}} contains 200 English items drawn from 200
distinct compounds, so compound identity cannot act as a shortcut: 50 per
condition, 100 aligned and 100 misleading. Each item carries three independent
human labels. The informed trio labelled 100 items and the blind trio the
other 100, with no item labelled by both.

\paragraph{Sentence length.} Length is measured in whitespace-separated tokens
with the \texttt{**\ldots**} markers removed. Overall the sentences run from 9
to 36 tokens, mean 19.6, median 19. Sentences from literal rows are a few tokens
longer than those from figurative rows (T1 18.2, T2 16.8, T3 21.4, T4 22.1),
which reflects a property of AdMIRe rather than of our sampling. Because length
covaries with the sentence sense rather than with alignment, it is balanced
across the aligned and misleading groups, which is the contrast the paper tests.

\paragraph{Label usage.} Table~\ref{tab:stats-labels} gives the distribution of
the 600 independent annotations. All four labels are used and the marginal
distribution is only mildly skewed, but within a condition it is severe by
construction: 87\% of annotations on figurative-sentence items fall on the
figurative side and 86\% on literal-sentence items fall on the literal side,
which confirms that the sentence fixes the reading as intended. Two consequences
follow. Per-class figures within a single condition rest on very small cells.
And because aligned-image labels concentrate at the ends of the scale, an
image-following shift has less room to appear than a shift toward the middle,
which bounds the effect the design could detect in the predicted direction.

\begin{table}[h]
\centering
\caption{Label usage: the 600 independent annotations, by condition. Each item
carries three labels from one trio.}
\label{tab:stats-labels}
\small
\setlength{\tabcolsep}{6pt}
\begin{tabular}{@{}lrrrrr@{}}
\toprule
 & \textbf{FF} & \textbf{WF} & \textbf{FL} & \textbf{LL} & \textbf{Total} \\
\midrule
T1 & 62 & 76 & 10 &  2 & 150 \\
T2 & 49 & 85 & 10 &  6 & 150 \\
T3 &  5 & 14 & 64 & 67 & 150 \\
T4 &  5 & 18 & 46 & 81 & 150 \\
\midrule
Aligned    &  54 &  99 & 74 & 73 & 300 \\
Misleading &  67 &  94 & 56 & 83 & 300 \\
\midrule
All & 121 & 193 & 130 & 156 & 600 \\
\bottomrule
\end{tabular}
\end{table}

\paragraph{Trio reliability.} Table~\ref{tab:stats-iaa} reports agreement on
the independent labels. The two trios are close and their intervals
overlap, so we do not treat the difference as a quality difference and report
both trios side by side throughout. The pooled row averages two disjoint trios
and is given for completeness only; it is not an estimate of any single trio's
reliability.

\begin{table}[h]
\centering
\caption{Inter-annotator agreement on independent labels.}
\vspace{2pt}
\label{tab:stats-iaa}
\small
\setlength{\tabcolsep}{5pt}
\begin{tabular}{@{}lrrrr@{}}
\toprule
\textbf{Trio} & \textbf{n} & \textbf{\% unanimous} & \textbf{Fleiss $\kappa$} &
\textbf{Mean Cohen $\kappa$} \\
\midrule
Informed & 100 & 63.0 & 0.65 & 0.66 \\
Blind    & 100 & 56.0 & 0.57 & 0.57 \\
\midrule
Pooled (not interpretable) & 200 & 59.5 & 0.62 & 0.62 \\
\bottomrule
\end{tabular}
\end{table}

\paragraph{Agreement by condition.} Table~\ref{tab:stats-iaa-cond} reports
Fleiss $\kappa$ per condition, separately by trio; throughout, $\Delta\kappa$
is aligned minus misleading, so a positive value is a drop under misleading
images. The informed trio is flat ($\Delta\kappa = -0.00$) and the blind
trio's point estimate moves by $0.10$ ($0.62 \to 0.52$), but neither shift is
statistically resolvable (Appendix~\ref{app:stats:tests}). Pooling the two
trios would average them and misrepresent both, so we report no pooled
aligned-versus-misleading gap. The per-condition cells hold roughly 25 items
each and are correspondingly noisy; the alignment rows, at 50 items, are the
ones the paper relies on.

\begin{table}[h]
\centering
\caption{Fleiss $\kappa$ per condition and per alignment, computed separately
within each trio and never pooled.}
\label{tab:stats-iaa-cond}
\small
\setlength{\tabcolsep}{5pt}
\begin{tabular}{@{}lrrrr@{}}
\toprule
 & \multicolumn{2}{c}{\textbf{Informed}} & \multicolumn{2}{c}{\textbf{Blind}} \\
 & \textbf{n} & \textbf{Fleiss $\kappa$} & \textbf{n} & \textbf{Fleiss $\kappa$} \\
\midrule
T1 (fig.\ sent., lit.\ img.)   & 26 & 0.50 & 24 & 0.30 \\
T2 (fig.\ sent., fig.\ img.)   & 26 & 0.52 & 24 & 0.44 \\
T3 (lit.\ sent., lit.\ img.)   & 24 & 0.63 & 26 & 0.53 \\
T4 (lit.\ sent., fig.\ img.)   & 24 & 0.62 & 26 & 0.41 \\
\midrule
Aligned    (T2 + T3)       & 50 & 0.65 & 50 & 0.62 \\
Misleading (T1 + T4)       & 50 & 0.65 & 50 & 0.52 \\
\midrule
$\Delta\kappa$ (aligned $-$ misleading) & & $-0.00$ & & $+0.10$ \\
\bottomrule
\end{tabular}
\end{table}

\subsection{Are the human contrasts resolvable?}
\label{app:stats:tests}

Two questions bear on the main paper. \textbf{(A)} Do the two trios annotate
equally reliably, or is the gap between them real? \textbf{(B)} Do annotators
agree less on misleading items than on aligned ones? We answer both with
procedures that differ in what they treat as an observation, and we never pool
the trios, which share no items.

\paragraph{Procedures.} The first resamples items with replacement 10{,}000
times, recomputes Fleiss $\kappa$ on each resample, and reports a 95\%
percentile interval on the difference. The second works at the item level: each
item is scored by the fraction of its three annotator pairs that assigned the
same label ($0$, $\tfrac{1}{3}$, $\tfrac{2}{3}$, $1$), and the 50 aligned items
are compared with the 50 misleading ones by a two-sided Welch $t$-test. Groups
are independent, since each compound appears in exactly one condition. Per-item
agreement is not chance-corrected, unlike $\kappa$, but the label distributions
are close across the two groups ($16/37/27/20$ aligned against $19/34/17/30$
misleading), so chance agreement is comparable and each trio is compared only
against itself. With $\mu_{\text{aln}}$ and $\mu_{\text{mis}}$ the mean
agreement under the two conditions, we test
$H_0\!: \mu_{\text{aln}} = \mu_{\text{mis}}$ against a two-sided alternative at
$\alpha = 0.05$.

\paragraph{(A) Trio effect.} Over all 100 items each, the informed trio
reaches Fleiss $\kappa = 0.65$ $[0.56, 0.74]$ and the blind trio $0.57$
$[0.46, 0.67]$; the difference is $+0.08$ $[-0.05, +0.22]$, $p = 0.22$. We
cannot reject that the two trios agree at the same rate, which is why we report
both trios rather than designating one as the reference.

\paragraph{(B) Alignment effect.} Table~\ref{tab:stats-humantests} reports the
aligned-versus-misleading contrast per trio under both procedures. Neither
trio shows a significant difference (blind $p = 0.27$, informed $p = 0.92$) and
every interval contains zero. The point estimates differ in kind, the informed
trio flat and the blind trio moving by roughly 8 agreement points in the
predicted direction, but the design cannot separate a shift of this size from noise: at 50 items per condition only differences of about $0.15$ or larger in
per-item agreement, and $0.20$ or larger in Fleiss $\kappa$, are resolvable.
Both figures are the half-width of the bootstrap interval on $\Delta$ under the
null. In the blind trio the movement comes from six items falling from
unanimous to two-of-three agreement, with outright three-way disagreement
unchanged at three items in both conditions, consistent with a single annotator
shifting by one class on a handful of items.

\begin{table}[H]
\centering
\caption{Aligned-versus-misleading contrasts, per trio. Left: Fleiss $\kappa$
with a 10{,}000-resample bootstrap interval. Right: mean per-item agreement with
a two-sided Welch $t$-test, $n = 50$ per group.}
\vspace{2pt}
\label{tab:stats-humantests}
\small
\setlength{\tabcolsep}{4pt}
\begin{tabular}{@{}lccccc@{}}
\toprule
 & \multicolumn{2}{c}{\textbf{Fleiss $\kappa$}} & \multicolumn{3}{c}{\textbf{Per-item agreement}} \\
\cmidrule(r){2-3}\cmidrule(l){4-6}
\textbf{Trio} & aln / mis & $\Delta\kappa$ [95\% CI] & aln / mis & $\Delta$ [95\% CI] & $p$ \\
\midrule
Informed & 0.65 / 0.65 & $-0.00$ $[-0.18, +0.18]$ & 0.740 / 0.747 & $-0.007$ $[-0.14, +0.13]$ & 0.92 \\
Blind    & 0.62 / 0.52 & $+0.10$ $[-0.10, +0.31]$ & 0.727 / 0.647 & $+0.080$ $[-0.06, +0.22]$ & 0.27 \\
\bottomrule
\end{tabular}
\end{table}

\paragraph{What this does and does not license.} These nulls do not establish
that human annotators are unaffected by misleading images; the blind trio's
interval admits a shift twice its point estimate. Nor do they establish an
effect. We therefore report the human contrast as unresolved, report both trios
side by side throughout, and treat neither as demonstrating robustness. The
nulls are themselves worth stating, since \citet{calderon2025alternative}
identify 50 to 100 annotated items as sufficient to reach a substitutability
verdict, yet at that size a condition-level contrast on the same annotators has
little power to detect whether the verdict would move under a different input.

\section{Change rates, all thirteen judges}
\label{app:change-full}

The six judges that pass the alt-test in no configuration, omitted from the
body, are GPT-4o \citep{openai2024gpt4o}, InternVL3.5-30B-A3B (IV-30B)
\citep{wang2025internvl35}, Kimi-VL-A3B-Instruct (K-VL)
\citep{kimiteam2025kimivl}, Qwen3-VL-32B (Q3-32B) \citep{qwenteam2025qwen3vl},
Qwen2.5-VL-7B (Q2.5-7B) \citep{bai2025qwen25vl} and Qwen2-VL-7B (Q2-7B)
\citep{wang2024qwen2vl}. They were run under exactly the same grid as the other seven.

\begin{table}[H]
\centering
\caption{Change rates for all thirteen judges, the full version of
Table~\ref{tab:change-main}. Columns are as in that table. Each judge contributes
the same number of cells to columns~1--3, so their means are also pooled rates;
column~4's denominator varies by judge, and pooling its cells gives 37\% rather
than the 34\% shown. $^{\circ}$ marks the six
judges that pass the alt-test in no configuration and are omitted from the
body table; every judge, in both groups, exceeds its own column~3 rate in both
image columns.}
\label{tab:change-full}
\footnotesize
\setlength{\tabcolsep}{5pt}
\begin{tabular}{@{}lrrrr@{}}
\toprule
& \multicolumn{3}{c}{\textbf{labels that change (\%)}} & \\
\cmidrule(lr){2-4}
\textbf{Judge}
& \makecell[r]{attaching an\\aligned image}
& \makecell[r]{attaching a\\misleading image}
& \makecell[r]{prompt edited,\\image fixed}
& \makecell[r]{of those that differ,\\\% toward the image} \\
\midrule
GPT-5.2    & 10.0 &  9.1 &  7.7 & 28 \\
GPT-4o$^{\circ}$  & 14.5 & 12.2 &  8.8 & 28 \\
G3.5-F     & 10.9 & 11.8 &  7.9 & 34 \\
G3.1-FL    & 14.5 & 13.0 &  8.4 & 12 \\
\midrule
Gm3-27B    & 17.9 & 19.2 & 10.6 & 30 \\
MiS-24B    & 21.2 & 16.9 &  9.1 & 21 \\
IV-30B$^{\circ}$  & 20.9 & 21.0 & 10.5 & 47 \\
K-VL$^{\circ}$    & 28.5 & 29.8 & 16.7 & 38 \\
Q3.6-35B   & 16.6 & 14.2 &  9.6 & 39 \\
Q3.6-27B   & 20.0 & 18.5 &  7.9 & 38 \\
Q3-32B$^{\circ}$  & 18.5 & 17.6 &  9.0 & 22 \\
Q2.5-7B$^{\circ}$ & 30.1 & 29.1 & 15.2 & 50 \\
Q2-7B$^{\circ}$   & 42.2 & 39.6 & 29.3 & 52 \\
\midrule
\textbf{mean (13)} & \textbf{20.5} & \textbf{19.4} & \textbf{11.6} & \textbf{34} \\
\bottomrule
\end{tabular}
\end{table}

\section{Full drift grid}
\label{app:drift-grid}

Table~\ref{tab:drift-appendix} gives the three shifts for every judge under all
four prompts, from which one row per judge and sentence type is selected as the
prompt most favourable to image-following; the two selections disagree for eight of the thirteen judges, and
both selected rows are marked in bold.

Three counts not reported in the body. On literal sentences the total shift is
negative for twelve of the thirteen judges, K-VL the exception at $+0.03$; the
aligned-to-misleading term runs back the other way for ten of them, with only
Q2-7B ($-0.13$) in the image-following direction and IV-30B and Q2.5-7B at zero.
Under Few-Shot + CoT, the guideline-consistent prompt rather than the selected
one, a single judge has that term in the predicted direction on both sentence
types, and 179 of the 200 majority-vote labels are unchanged when the image is replaced.

Two patterns the selection hides. Text-only to aligned exceeds aligned to
misleading in magnitude in most cells, so a study contrasting only text-only
against the misleading image would attribute to the manipulation what the
aligned arm shows to be an effect of having any image at all. And aligned to
misleading is unstable across prompts in a way the total is not: on literal
sentences K-VL ranges from $+0.01$ under Few-Shot + CoT to $+0.32$ under CoT, and
Q2.5-7B from $-0.32$ under Few-Shot to $+0.36$ under CoT, with no consistent
ordering of prompts across judges.

\paragraph{Intervals on the content term.} For each judge and sentence type we
hold fixed that selected prompt, resample the 200
compounds with replacement 10{,}000 times, and take a 95\% percentile interval
on the mean aligned-to-misleading shift. Twenty-three of the 26 intervals
contain zero and all lie within $[-0.42, +0.25]$; of the three exceptions, two
run in the image-following direction and one against it. Since the selection
takes a maximum over four prompts per cell, three exceptions out of 26 is close
to what selection alone would produce. Repeating the procedure under Few-Shot +
CoT throughout leaves 22 of 26 intervals containing zero, and all four
exceptions fall on literal sentences in the anti-image-following direction.

\begingroup
\scriptsize
\setlength{\tabcolsep}{4pt}
\begin{longtable}{@{}ll rrr rrr@{}}
\caption{Drift decomposition for all thirteen judges under each prompt. Mean
ordinal shift (FF{=}0..LL{=}3); \follow{blue} marks the predicted sign,
positive on figurative and negative on literal sentences. \textbf{Bold} marks
the selected cell, per judge and sentence type. Judges grouped by family.}
\label{tab:drift-appendix} \\
\toprule
& & \multicolumn{3}{c}{\textbf{Figurative sentences}} & \multicolumn{3}{c}{\textbf{Literal sentences}} \\
\cmidrule(lr){3-5}\cmidrule(l){6-8}
\textbf{Judge} & \textbf{Prompt}
& \makecell[r]{text-only\\$\to$ aligned} & \makecell[r]{aligned\\$\to$ misleading} & \makecell[r]{text-only\\$\to$ misleading}
& \makecell[r]{text-only\\$\to$ aligned} & \makecell[r]{aligned\\$\to$ misleading} & \makecell[r]{text-only\\$\to$ misleading} \\
\midrule
\endfirsthead

\multicolumn{8}{@{}l}{\emph{Table~\ref{tab:drift-appendix} continued}}\\
\toprule
& & \multicolumn{3}{c}{\textbf{Figurative sentences}} & \multicolumn{3}{c}{\textbf{Literal sentences}} \\
\cmidrule(lr){3-5}\cmidrule(l){6-8}
\textbf{Judge} & \textbf{Prompt}
& \makecell[r]{text-only\\$\to$ aligned} & \makecell[r]{aligned\\$\to$ misleading} & \makecell[r]{text-only\\$\to$ misleading}
& \makecell[r]{text-only\\$\to$ aligned} & \makecell[r]{aligned\\$\to$ misleading} & \makecell[r]{text-only\\$\to$ misleading} \\
\midrule
\endhead

\midrule
\multicolumn{8}{r@{}}{\emph{Continued on next page}} \\
\endfoot

\bottomrule
\endlastfoot

\multirow{4}{*}{GPT-5.2}
& Zero-Shot& $-0.04$ & \follow{$+0.01$} & $-0.03$ & $-0.07$ & $+0.04$ & \follow{$-0.03$} \\*
& Few-Shot & \textbf{$+0.04$} & \textbf{$-0.01$} & \textbf{\follow{$+0.03$}} & \textbf{$-0.18$} & \textbf{$+0.07$} & \textbf{\follow{$-0.11$}} \\*
& CoT  & $\phantom{+}0.00$ & $-0.07$ & $-0.07$ & $-0.10$ & $+0.05$ & \follow{$-0.05$} \\*
& Few-Shot + CoT  & $-0.02$ & $-0.04$ & $-0.06$ & $-0.09$ & $+0.05$ & \follow{$-0.04$} \\
\midrule
\multirow{4}{*}{GPT-4o}
& Zero-Shot& $+0.04$ & $-0.17$ & $-0.13$ & $-0.18$ & $+0.19$ & $+0.01$ \\*
& Few-Shot & \textbf{$+0.07$} & \textbf{$-0.07$} & \textbf{$\phantom{+}0.00$} & $-0.15$ & $+0.06$ & \follow{$-0.09$} \\*
& CoT  & $+0.03$ & $-0.07$ & $-0.04$ & $+0.01$ & $+0.02$ & $+0.03$ \\*
& Few-Shot + CoT  & $-0.03$ & $-0.01$ & $-0.04$ & \textbf{$-0.12$} & \textbf{$+0.02$} & \textbf{\follow{$-0.10$}} \\
\midrule\midrule
\multirow{4}{*}{G3.5-F}
& Zero-Shot& $-0.02$ & $-0.04$ & $-0.06$ & $-0.04$ & $+0.09$ & $+0.05$ \\*
& Few-Shot & \textbf{$+0.09$} & \textbf{$-0.04$} & \textbf{\follow{$+0.05$}} & \textbf{$-0.07$} & \textbf{$+0.01$} & \textbf{\follow{$-0.06$}} \\*
& CoT  & $+0.03$ & $-0.02$ & \follow{$+0.01$} & $-0.01$ & $+0.08$ & $+0.07$ \\*
& Few-Shot + CoT  & $-0.02$ & $-0.01$ & $-0.03$ & $-0.07$ & $+0.08$ & $+0.01$ \\
\midrule
\multirow{4}{*}{G3.1-FL}
& Zero-Shot& $-0.05$ & $-0.04$ & $-0.09$ & \textbf{$-0.23$} & \textbf{$+0.15$} & \textbf{\follow{$-0.08$}} \\*
& Few-Shot & $-0.05$ & $-0.06$ & $-0.11$ & $-0.08$ & $+0.10$ & $+0.02$ \\*
& CoT  & \textbf{$+0.01$} & \textbf{$-0.06$} & \textbf{$-0.05$} & $-0.17$ & $+0.26$ & $+0.09$ \\*
& Few-Shot + CoT  & $-0.02$ & $-0.05$ & $-0.07$ & $-0.12$ & $+0.21$ & $+0.09$ \\
\midrule\midrule
\multirow{4}{*}{Gm3-27B}
& Zero-Shot& \textbf{$+0.03$} & \textbf{\follow{$+0.02$}} & \textbf{\follow{$+0.05$}} & \textbf{$-0.14$} & \textbf{$+0.01$} & \textbf{\follow{$-0.13$}} \\*
& Few-Shot & $+0.08$ & $-0.11$ & $-0.03$ & $+0.01$ & $+0.19$ & $+0.20$ \\*
& CoT  & $-0.02$ & $-0.02$ & $-0.04$ & $+0.04$ & $+0.03$ & $+0.07$ \\*
& Few-Shot + CoT  & $+0.08$ & $-0.07$ & \follow{$+0.01$} & $-0.02$ & $+0.14$ & $+0.12$ \\
\midrule\midrule
\multirow{4}{*}{MiS-24B}
& Zero-Shot& \textbf{$+0.08$} & \textbf{$-0.03$} & \textbf{\follow{$+0.05$}} & \textbf{$-0.39$} & \textbf{$+0.14$} & \textbf{\follow{$-0.25$}} \\*
& Few-Shot & $-0.04$ & $-0.10$ & $-0.14$ & $-0.22$ & $+0.21$ & \follow{$-0.01$} \\*
& CoT  & $-0.12$ & $\phantom{+}0.00$ & $-0.12$ & $-0.16$ & $+0.25$ & $+0.09$ \\*
& Few-Shot + CoT  & $+0.02$ & $-0.01$ & \follow{$+0.01$} & $-0.08$ & $+0.09$ & $+0.01$ \\
\midrule\midrule
\multirow{4}{*}{IV-30B}
& Zero-Shot& $+0.03$ & $-0.12$ & $-0.09$ & $-0.11$ & $-0.02$ & \follow{$-0.13$} \\*
& Few-Shot & $-0.11$ & $-0.04$ & $-0.16$ & $+0.03$ & \follow{$-0.13$} & \follow{$-0.10$} \\*
& CoT  & $-0.04$ & $-0.01$ & $-0.05$ & \textbf{$-0.33$} & \textbf{$\phantom{+}0.00$} & \textbf{\follow{$-0.33$}} \\*
& Few-Shot + CoT  & \textbf{$-0.01$} & \textbf{\follow{$+0.01$}} & \textbf{$\phantom{+}0.00$} & $+0.16$ & $-0.05$ & $+0.11$ \\
\midrule\midrule
\multirow{4}{*}{K-VL}
& Zero-Shot& $+0.06$ & $-0.05$ & \follow{$+0.01$} & $-0.03$ & $+0.19$ & $+0.16$ \\*
& Few-Shot & \textbf{$+0.20$} & \textbf{\follow{$+0.02$}} & \textbf{\follow{$+0.22$}} & $+0.16$ & $+0.14$ & $+0.30$ \\*
& CoT  & $-0.22$ & $-0.02$ & $-0.24$ & \textbf{$-0.29$} & \textbf{$+0.32$} & \textbf{$+0.03$} \\*
& Few-Shot + CoT  & $+0.16$ & $-0.07$ & \follow{$+0.08$} & $+0.16$ & $+0.01$ & $+0.17$ \\
\midrule\midrule
\multirow{4}{*}{Q3.6-35B}
& Zero-Shot& $-0.03$ & $-0.05$ & $-0.08$ & \textbf{$-0.27$} & \textbf{$+0.11$} & \textbf{\follow{$-0.16$}} \\*
& Few-Shot & $+0.06$ & $-0.03$ & \follow{$+0.03$} & $+0.02$ & $+0.01$ & $+0.03$ \\*
& CoT  & $-0.03$ & \follow{$+0.02$} & $-0.01$ & $-0.13$ & $+0.05$ & \follow{$-0.08$} \\*
& Few-Shot + CoT  & \textbf{$+0.07$} & \textbf{$-0.02$} & \textbf{\follow{$+0.05$}} & $+0.03$ & $+0.01$ & $+0.04$ \\
\midrule
\multirow{4}{*}{Q3.6-27B}
& Zero-Shot& \textbf{$+0.26$} & \textbf{$\phantom{+}0.00$} & \textbf{\follow{$+0.26$}} & \textbf{$-0.26$} & \textbf{$+0.04$} & \textbf{\follow{$-0.22$}} \\*
& Few-Shot & $+0.03$ & $-0.03$ & $\phantom{+}0.00$ & $-0.16$ & $+0.10$ & \follow{$-0.06$} \\*
& CoT  & $-0.10$ & $-0.01$ & $-0.11$ & $+0.07$ & $+0.05$ & $+0.12$ \\*
& Few-Shot + CoT  & $-0.08$ & \follow{$+0.06$} & $-0.02$ & $-0.16$ & $+0.08$ & \follow{$-0.08$} \\
\midrule
\multirow{4}{*}{Q3-32B}
& Zero-Shot& $-0.13$ & $-0.26$ & $-0.39$ & \textbf{$-0.44$} & \textbf{$+0.12$} & \textbf{\follow{$-0.32$}} \\*
& Few-Shot & \textbf{$+0.14$} & \textbf{$-0.11$} & \textbf{\follow{$+0.03$}} & $-0.17$ & $+0.14$ & \follow{$-0.03$} \\*
& CoT  & $+0.02$ & $-0.07$ & $-0.05$ & $-0.25$ & $+0.09$ & \follow{$-0.16$} \\*
& Few-Shot + CoT  & $\phantom{+}0.00$ & \follow{$+0.02$} & \follow{$+0.02$} & $-0.07$ & $+0.04$ & \follow{$-0.03$} \\
\midrule
\multirow{4}{*}{Q2.5-7B}
& Zero-Shot& $-0.09$ & $-0.04$ & $-0.13$ & \textbf{$-0.23$} & \textbf{$\phantom{+}0.00$} & \textbf{\follow{$-0.23$}} \\*
& Few-Shot & \textbf{$+0.06$} & \textbf{$-0.07$} & \textbf{$-0.01$} & $+0.26$ & \follow{$-0.32$} & \follow{$-0.06$} \\*
& CoT  & $-0.18$ & $-0.08$ & $-0.26$ & $-0.30$ & $+0.36$ & $+0.06$ \\*
& Few-Shot + CoT  & $-0.26$ & \follow{$+0.10$} & $-0.16$ & $\phantom{+}0.00$ & $+0.06$ & $+0.06$ \\
\midrule
\multirow{4}{*}{Q2-7B}
& Zero-Shot& $+0.03$ & \follow{$+0.05$} & \follow{$+0.08$} & $+0.32$ & \follow{$-0.11$} & $+0.21$ \\*
& Few-Shot & $\phantom{+}0.00$ & $-0.20$ & $-0.20$ & \textbf{$-0.17$} & \textbf{\follow{$-0.13$}} & \textbf{\follow{$-0.30$}} \\*
& CoT  & \textbf{$+0.24$} & \textbf{$-0.01$} & \textbf{\follow{$+0.23$}} & $+0.28$ & \follow{$-0.08$} & $+0.20$ \\*
& Few-Shot + CoT  & $-0.07$ & $-0.03$ & $-0.10$ & $+0.08$ & \follow{$-0.18$} & \follow{$-0.10$} \\
\end{longtable}
\endgroup

\paragraph{Suppression, not indifference.} For each judge we count the items
whose label changes between text-only and the aligned image ($S$) and the
subset moving toward the sense that image depicts ($F$). A judge ignoring the
image gives $F/S \approx 0.5$; below that, changes run counter to the image.
Table~\ref{tab:nd} reports the ratio under Few-Shot + CoT. One judge exceeds
$0.5$ (Q2.5-7B, $0.60$) and six fall below $0.4$, so the modal pattern is
active suppression rather than inattention.

\begin{table}[h]
\centering
\caption{Direction of label changes under the aligned image, Few-Shot + CoT.
$S$ is the number of items whose label changes from text-only; $F$ the subset
moving toward that image's sense.}
\label{tab:nd}
\small
\begin{tabular}{@{}lrrr@{}}
\toprule
\textbf{Judge} & \textbf{S} & \textbf{F} & \textbf{F/S} \\
\midrule
Q2.5-7B  & 65 & 39 & 0.60 \\
K-VL     & 86 & 43 & 0.50 \\
GPT-4o   & 19 &  9 & 0.47 \\
Q3.6-27B & 32 & 15 & 0.47 \\
IV-30B   & 52 & 24 & 0.46 \\
Q2-7B    & 76 & 35 & 0.46 \\
Q3.6-35B & 22 &  8 & 0.36 \\
Q3-32B   & 23 &  8 & 0.35 \\
MiS-24B  & 33 & 10 & 0.30 \\
Gm3-27B  & 30 &  9 & 0.30 \\
G3.5-F   & 17 &  5 & 0.29 \\
G3.1-FL  & 21 &  6 & 0.29 \\
GPT-5.2  & 14 &  2 & 0.14 \\
\bottomrule
\end{tabular}
\end{table}

\section{Alt-test grid}
\label{app:alttest}

Table~\ref{tab:alt-test-appendix-full} gives the winning rate (WR) for every judge,
prompt, trio, condition and slack, under both scoring functions.

No cell passes on the informed trio at either slack, under either scoring
function, in any condition or prompt; the informed columns are uniformly zero
apart from a single $0.33$ for Q3.6-27B, one annotator of three and below the
threshold. On the blind trio the pooled condition carries every pass: two
judges clear at $\varepsilon = 0.15$, G3.5-F and Gm3-27B, and seven at $0.20$,
adding GPT-5.2, G3.1-FL, MiS-24B, Q3.6-35B and Q3.6-27B. Six judges never pass
in any cell, GPT-4o, IV-30B, K-VL, Q3-32B, Q2.5-7B and Q2-7B, so the slack
separates a middle band rather than shifting the whole field.

The two scoring functions agree: the set of judges passing the pooled condition is
identical under exact match (ACC) and ordinal distance (MAE) at both slacks, and
where the two differ no judge's status changes. Nothing in the argument depends on whether agreement is
scored by exact match or by ordinal distance.

The condition slices are unstable in both directions at $\varepsilon = 0.20$.
G3.1-FL under Zero-Shot passes on the aligned half and fails on the hundred
items containing it, while Gm3-27B under Zero-Shot passes on the misleading half
and not the aligned one. Since more data should make a real advantage easier to
detect rather than harder, we read these as noise on fifty-item slices rather
than as a condition effect. The direction of the trio difference, by contrast,
has a mechanism: advantage probability runs higher on the blind trio for the
same judge because its annotators agree less with one another, which lowers the
bar a withheld human sets. A judge looks more substitutable against a noisier
trio, which is a property of the procedure rather than a fault in it.

\begingroup
\tiny
\setlength{\tabcolsep}{1.6pt}
\begin{longtable}{@{}ll rr rr rr rr rr rr rr rr rr rr rr rr@{}}
\caption{Alt-test winning rate per judge and prompt on both trio, at
$\varepsilon \in \{0.15, 0.20\}$ under two scoring functions: \textbf{ACC},
exact match, and \textbf{MAE}, negative mean absolute error on the ordinal codes
FF{=}0 to LL{=}3. Green marks a pass, WR $\geq 0.5$. Conditions are aligned
(T2{+}T3), misleading (T1{+}T4) and all (T1--T4); single T-types are omitted, as
$n \approx 25$ falls below the recommended minimum per annotator.}
\label{tab:alt-test-appendix-full} \\
\toprule
 &  & \multicolumn{12}{c}{\textbf{Informed trio}} & \multicolumn{12}{c}{\textbf{Blind trio}} \\
\cmidrule(lr){3-14} \cmidrule(lr){15-26}
 &  & \multicolumn{4}{c}{aligned} & \multicolumn{4}{c}{misleading} & \multicolumn{4}{c}{all} & \multicolumn{4}{c}{aligned} & \multicolumn{4}{c}{misleading} & \multicolumn{4}{c}{all} \\
\cmidrule(lr){3-6} \cmidrule(lr){7-10} \cmidrule(lr){11-14} \cmidrule(lr){15-18} \cmidrule(lr){19-22} \cmidrule(lr){23-26}
 &  & \multicolumn{2}{c}{ACC} & \multicolumn{2}{c}{MAE} & \multicolumn{2}{c}{ACC} & \multicolumn{2}{c}{MAE} & \multicolumn{2}{c}{ACC} & \multicolumn{2}{c}{MAE} & \multicolumn{2}{c}{ACC} & \multicolumn{2}{c}{MAE} & \multicolumn{2}{c}{ACC} & \multicolumn{2}{c}{MAE} & \multicolumn{2}{c}{ACC} & \multicolumn{2}{c}{MAE} \\
\cmidrule(lr){3-4} \cmidrule(lr){5-6} \cmidrule(lr){7-8} \cmidrule(lr){9-10} \cmidrule(lr){11-12} \cmidrule(lr){13-14} \cmidrule(lr){15-16} \cmidrule(lr){17-18} \cmidrule(lr){19-20} \cmidrule(lr){21-22} \cmidrule(lr){23-24} \cmidrule(lr){25-26}
\textbf{Judge} & \textbf{Prompt} & .15 & .20 & .15 & .20 & .15 & .20 & .15 & .20 & .15 & .20 & .15 & .20 & .15 & .20 & .15 & .20 & .15 & .20 & .15 & .20 & .15 & .20 & .15 & .20 \\
\midrule
\endfirsthead

\multicolumn{26}{@{}l}{\emph{Table~\ref{tab:alt-test-appendix-full} continued}}\\
\toprule
 &  & \multicolumn{12}{c}{\textbf{Informed trio}} & \multicolumn{12}{c}{\textbf{Blind trio}} \\
\cmidrule(lr){3-14} \cmidrule(lr){15-26}
 &  & \multicolumn{4}{c}{aligned} & \multicolumn{4}{c}{misleading} & \multicolumn{4}{c}{all} & \multicolumn{4}{c}{aligned} & \multicolumn{4}{c}{misleading} & \multicolumn{4}{c}{all} \\
\cmidrule(lr){3-6} \cmidrule(lr){7-10} \cmidrule(lr){11-14} \cmidrule(lr){15-18} \cmidrule(lr){19-22} \cmidrule(lr){23-26}
 &  & \multicolumn{2}{c}{ACC} & \multicolumn{2}{c}{MAE} & \multicolumn{2}{c}{ACC} & \multicolumn{2}{c}{MAE} & \multicolumn{2}{c}{ACC} & \multicolumn{2}{c}{MAE} & \multicolumn{2}{c}{ACC} & \multicolumn{2}{c}{MAE} & \multicolumn{2}{c}{ACC} & \multicolumn{2}{c}{MAE} & \multicolumn{2}{c}{ACC} & \multicolumn{2}{c}{MAE} \\
\cmidrule(lr){3-4} \cmidrule(lr){5-6} \cmidrule(lr){7-8} \cmidrule(lr){9-10} \cmidrule(lr){11-12} \cmidrule(lr){13-14} \cmidrule(lr){15-16} \cmidrule(lr){17-18} \cmidrule(lr){19-20} \cmidrule(lr){21-22} \cmidrule(lr){23-24} \cmidrule(lr){25-26}
\textbf{Judge} & \textbf{Prompt} & .15 & .20 & .15 & .20 & .15 & .20 & .15 & .20 & .15 & .20 & .15 & .20 & .15 & .20 & .15 & .20 & .15 & .20 & .15 & .20 & .15 & .20 & .15 & .20 \\
\midrule
\endhead

\midrule
\multicolumn{26}{r@{}}{\emph{Continued on next page}} \\
\endfoot

\bottomrule
\endlastfoot

\multirow{4}{*}{GPT-5.2} & Zero-Shot & .00 & .00 & .00 & .00 & .00 & .00 & .00 & .00 & .00 & .00 & .00 & .00 & .00 & .00 & .00 & .00 & .00 & .00 & .00 & .00 & .00 & .00 & .00 & .00 \\*
 & Few-Shot & .00 & .00 & .00 & .00 & .00 & .00 & .00 & .00 & .00 & .00 & .00 & .00 & .00 & .00 & .00 & .00 & .00 & .00 & .00 & .00 & .00 & .33 & .00 & .33 \\*
 & CoT & .00 & .00 & .00 & .00 & .00 & .00 & .00 & .00 & .00 & .00 & .00 & .00 & .00 & .00 & .00 & .00 & .00 & .00 & .00 & .00 & .00 & .33 & .00 & .33 \\*
 & Few-Shot + CoT & .00 & .00 & .00 & .00 & .00 & .00 & .00 & .00 & .00 & .00 & .00 & .00 & .00 & .00 & .00 & .00 & .33 & \pass{1.0} & .33 & \pass{1.0} & .33 & \pass{1.0} & .33 & \pass{1.0} \\
\midrule
\multirow{4}{*}{GPT-4o} & Zero-Shot & .00 & .00 & .00 & .00 & .00 & .00 & .00 & .00 & .00 & .00 & .00 & .00 & .00 & .00 & .00 & .00 & .00 & .00 & .00 & .00 & .00 & .00 & .00 & .00 \\*
 & Few-Shot & .00 & .00 & .00 & .00 & .00 & .00 & .00 & .00 & .00 & .00 & .00 & .00 & .00 & .00 & .00 & .00 & .00 & .00 & .00 & .00 & .00 & .00 & .00 & .00 \\*
 & CoT & .00 & .00 & .00 & .00 & .00 & .00 & .00 & .00 & .00 & .00 & .00 & .00 & .00 & .00 & .00 & .00 & .00 & .00 & .00 & .00 & .00 & .00 & .00 & .00 \\*
 & Few-Shot + CoT & .00 & .00 & .00 & .00 & .00 & .00 & .00 & .00 & .00 & .00 & .00 & .00 & .00 & .00 & .00 & .00 & .00 & .00 & .00 & .00 & .00 & .33 & .00 & .00 \\
\midrule\midrule
\multirow{4}{*}{G3.5-F} & Zero-Shot & .00 & .00 & .00 & .00 & .00 & .00 & .00 & .00 & .00 & .00 & .00 & .00 & .00 & .00 & .00 & .00 & .00 & \pass{.67} & .00 & .33 & .33 & \pass{1.0} & .33 & \pass{1.0} \\*
 & Few-Shot & .00 & .00 & .00 & .00 & .00 & .00 & .00 & .00 & .00 & .00 & .00 & .00 & .00 & .00 & .00 & .00 & .00 & .00 & .00 & .00 & .00 & \pass{1.0} & .00 & \pass{1.0} \\*
 & CoT & .00 & .00 & .00 & .00 & .00 & .00 & .00 & .00 & .00 & .00 & .00 & .00 & .00 & .00 & .00 & .00 & .00 & .33 & .00 & .33 & .33 & \pass{1.0} & .33 & \pass{1.0} \\*
 & Few-Shot + CoT & .00 & .00 & .00 & .00 & .00 & .00 & .00 & .00 & .00 & .00 & .00 & .00 & .00 & \pass{1.0} & \pass{1.0} & \pass{1.0} & .33 & \pass{1.0} & .33 & \pass{1.0} & \pass{1.0} & \pass{1.0} & \pass{1.0} & \pass{1.0} \\
\midrule
\multirow{4}{*}{G3.1-FL} & Zero-Shot & .00 & .00 & .00 & .00 & .00 & .00 & .00 & .00 & .00 & .00 & .00 & .00 & .00 & \pass{1.0} & .00 & \pass{1.0} & .00 & .00 & .00 & .00 & .00 & .00 & .00 & .00 \\*
 & Few-Shot & .00 & .00 & .00 & .00 & .00 & .00 & .00 & .00 & .00 & .00 & .00 & .00 & .00 & \pass{1.0} & .00 & \pass{1.0} & .00 & .00 & .00 & .00 & .00 & \pass{1.0} & .00 & \pass{1.0} \\*
 & CoT & .00 & .00 & .00 & .00 & .00 & .00 & .00 & .00 & .00 & .00 & .00 & .00 & .00 & .00 & .00 & .00 & .00 & .00 & .00 & .00 & .00 & .00 & .00 & .00 \\*
 & Few-Shot + CoT & .00 & .00 & .00 & .00 & .00 & .00 & .00 & .00 & .00 & .00 & .00 & .00 & .00 & \pass{1.0} & .00 & \pass{1.0} & .00 & .33 & .00 & .00 & .33 & \pass{1.0} & .33 & \pass{1.0} \\
\midrule\midrule
\multirow{4}{*}{Gm3-27B} & Zero-Shot & .00 & .00 & .00 & .00 & .00 & .00 & .00 & .00 & .00 & .00 & .00 & .00 & .00 & .00 & .00 & .00 & .00 & \pass{1.0} & .33 & \pass{1.0} & .00 & \pass{.67} & .00 & \pass{1.0} \\*
 & Few-Shot & .00 & .00 & .00 & .00 & .00 & .00 & .00 & .00 & .00 & .00 & .00 & .00 & .00 & \pass{1.0} & .00 & \pass{1.0} & .00 & \pass{.67} & .00 & .33 & \pass{.67} & \pass{1.0} & \pass{.67} & \pass{1.0} \\*
 & CoT & .00 & .00 & .00 & .00 & .00 & .00 & .00 & .00 & .00 & .00 & .00 & .00 & .00 & .00 & .00 & .00 & .00 & .00 & .00 & .00 & .00 & .00 & .00 & .00 \\*
 & Few-Shot + CoT & .00 & .00 & .00 & .00 & .00 & .00 & .00 & .00 & .00 & .00 & .00 & .00 & .00 & .00 & .00 & .00 & .00 & .00 & .00 & .00 & .00 & .00 & .00 & .00 \\
\midrule\midrule
\multirow{4}{*}{MiS-24B} & Zero-Shot & .00 & .00 & .00 & .00 & .00 & .00 & .00 & .00 & .00 & .00 & .00 & .00 & .00 & .00 & .00 & .00 & .33 & .33 & .00 & .33 & .00 & .00 & .00 & \pass{1.0} \\*
 & Few-Shot & .00 & .00 & .00 & .00 & .00 & .00 & .00 & .00 & .00 & .00 & .00 & .00 & .00 & .00 & .00 & .00 & .00 & .00 & .00 & .00 & .00 & .00 & .00 & .00 \\*
 & CoT & .00 & .00 & .00 & .00 & .00 & .00 & .00 & .00 & .00 & .00 & .00 & .00 & .00 & .00 & .00 & .00 & .00 & .00 & .00 & .00 & .00 & .00 & .00 & .00 \\*
 & Few-Shot + CoT & .00 & .00 & .00 & .00 & .00 & .00 & .00 & .00 & .00 & .00 & .00 & .00 & .00 & .00 & .00 & .00 & .33 & .33 & .00 & .33 & .33 & \pass{1.0} & .00 & \pass{1.0} \\
\midrule\midrule
\multirow{4}{*}{IV-30B} & Zero-Shot & .00 & .00 & .00 & .00 & .00 & .00 & .00 & .00 & .00 & .00 & .00 & .00 & .00 & .00 & .00 & .00 & .00 & .00 & .00 & .00 & .00 & .00 & .00 & .00 \\*
 & Few-Shot & .00 & .00 & .00 & .00 & .00 & .00 & .00 & .00 & .00 & .00 & .00 & .00 & .00 & .00 & .00 & .00 & .00 & .00 & .00 & .00 & .00 & .00 & .00 & .00 \\*
 & CoT & .00 & .00 & .00 & .00 & .00 & .00 & .00 & .00 & .00 & .00 & .00 & .00 & .00 & .00 & .00 & .00 & .00 & .00 & .00 & .00 & .00 & .00 & .00 & .00 \\*
 & Few-Shot + CoT & .00 & .00 & .00 & .00 & .00 & .00 & .00 & .00 & .00 & .00 & .00 & .00 & .00 & .00 & .00 & .00 & .00 & .00 & .00 & .00 & .00 & .00 & .00 & .00 \\
\midrule\midrule
\multirow{4}{*}{K-VL} & Zero-Shot & .00 & .00 & .00 & .00 & .00 & .00 & .00 & .00 & .00 & .00 & .00 & .00 & .00 & .00 & .00 & .00 & .00 & .00 & .00 & .00 & .00 & .00 & .00 & .00 \\*
 & Few-Shot & .00 & .00 & .00 & .00 & .00 & .00 & .00 & .00 & .00 & .00 & .00 & .00 & .00 & .00 & .00 & .00 & .00 & .00 & .00 & .00 & .00 & .00 & .00 & .00 \\*
 & CoT & .00 & .00 & .00 & .00 & .00 & .00 & .00 & .00 & .00 & .00 & .00 & .00 & .00 & .00 & .00 & .00 & .00 & .00 & .00 & .00 & .00 & .00 & .00 & .00 \\*
 & Few-Shot + CoT & .00 & .00 & .00 & .00 & .00 & .00 & .00 & .00 & .00 & .00 & .00 & .00 & .00 & .00 & .00 & .00 & .00 & .00 & .00 & .00 & .00 & .00 & .00 & .00 \\
\midrule\midrule
\multirow{4}{*}{Q3.6-35B} & Zero-Shot & .00 & .00 & .00 & .00 & .00 & .00 & .00 & .00 & .00 & .00 & .00 & .00 & .00 & .00 & .00 & .00 & .00 & .00 & .00 & .00 & .00 & .00 & .00 & .00 \\*
 & Few-Shot & .00 & .00 & .00 & .00 & .00 & .00 & .00 & .00 & .00 & .00 & .00 & .00 & .00 & .00 & .00 & .00 & .00 & .33 & .00 & .33 & .00 & \pass{.67} & .00 & \pass{1.0} \\*
 & CoT & .00 & .00 & .00 & .00 & .00 & .00 & .00 & .00 & .00 & .00 & .00 & .00 & .00 & .00 & .00 & \pass{1.0} & .00 & .00 & .00 & .00 & .00 & \pass{.67} & .00 & \pass{.67} \\*
 & Few-Shot + CoT & .00 & .00 & .00 & .00 & .00 & .00 & .00 & .00 & .00 & .00 & .00 & .00 & .00 & \pass{1.0} & .00 & \pass{1.0} & .00 & .33 & .00 & .33 & .33 & \pass{1.0} & .33 & \pass{1.0} \\
\midrule
\multirow{4}{*}{Q3.6-27B} & Zero-Shot & .00 & .00 & .00 & .00 & .00 & .00 & .00 & .00 & .00 & .00 & .00 & .00 & .00 & .00 & .00 & .00 & .00 & .00 & .00 & .00 & .00 & .00 & .00 & .00 \\*
 & Few-Shot & .00 & .00 & .00 & .00 & .00 & .00 & .00 & .00 & .00 & .00 & .00 & .00 & .00 & .00 & .00 & \pass{1.0} & .00 & .00 & .00 & .00 & .00 & .00 & .00 & \pass{1.0} \\*
 & CoT & .00 & .33 & .00 & .00 & .00 & .00 & .00 & .00 & .00 & .00 & .00 & .00 & .00 & .00 & .00 & .00 & .00 & .33 & .00 & .00 & .33 & \pass{1.0} & .00 & \pass{.67} \\*
 & Few-Shot + CoT & .00 & .00 & .00 & .00 & .00 & .00 & .00 & .00 & .00 & .00 & .00 & .00 & .00 & .00 & .00 & .00 & .00 & .00 & .00 & .00 & .00 & \pass{1.0} & .00 & \pass{1.0} \\
\midrule\midrule
\multirow{4}{*}{Q3-32B} & Zero-Shot & .00 & .00 & .00 & .00 & .00 & .00 & .00 & .00 & .00 & .00 & .00 & .00 & .00 & .00 & .00 & .00 & .00 & .00 & .00 & .00 & .00 & .00 & .00 & .00 \\*
 & Few-Shot & .00 & .00 & .00 & .00 & .00 & .00 & .00 & .00 & .00 & .00 & .00 & .00 & .00 & .00 & .00 & .00 & .00 & .00 & .00 & .00 & .00 & .00 & .00 & .00 \\*
 & CoT & .00 & .00 & .00 & .00 & .00 & .00 & .00 & .00 & .00 & .00 & .00 & .00 & .00 & .00 & .00 & .00 & .00 & .00 & .00 & .00 & .00 & .00 & .00 & .00 \\*
 & Few-Shot + CoT & .00 & .00 & .00 & .00 & .00 & .00 & .00 & .00 & .00 & .00 & .00 & .00 & .00 & .00 & .00 & .00 & .00 & .33 & .00 & .00 & .00 & .33 & .00 & .33 \\
\midrule
\multirow{4}{*}{Q2.5-7B} & Zero-Shot & .00 & .00 & .00 & .00 & .00 & .00 & .00 & .00 & .00 & .00 & .00 & .00 & .00 & .00 & .00 & .00 & .00 & .00 & .00 & .00 & .00 & .00 & .00 & .00 \\*
 & Few-Shot & .00 & .00 & .00 & .00 & .00 & .00 & .00 & .00 & .00 & .00 & .00 & .00 & .00 & .00 & .00 & .00 & .00 & .00 & .00 & .00 & .00 & .00 & .00 & .00 \\*
 & CoT & .00 & .00 & .00 & .00 & .00 & .00 & .00 & .00 & .00 & .00 & .00 & .00 & .00 & .00 & .00 & .00 & .00 & .00 & .00 & .00 & .00 & .00 & .00 & .00 \\*
 & Few-Shot + CoT & .00 & .00 & .00 & .00 & .00 & .00 & .00 & .00 & .00 & .00 & .00 & .00 & .00 & .00 & .00 & .00 & .00 & .00 & .00 & .00 & .00 & .00 & .00 & .00 \\
\midrule
\multirow{4}{*}{Q2-7B} & Zero-Shot & .00 & .00 & .00 & .00 & .00 & .00 & .00 & .00 & .00 & .00 & .00 & .00 & .00 & .00 & .00 & .00 & .00 & .00 & .00 & .00 & .00 & .00 & .00 & .00 \\*
 & Few-Shot & .00 & .00 & .00 & .00 & .00 & .00 & .00 & .00 & .00 & .00 & .00 & .00 & .00 & .00 & .00 & .00 & .00 & .00 & .00 & .00 & .00 & .00 & .00 & .00 \\*
 & CoT & .00 & .00 & .00 & .00 & .00 & .00 & .00 & .00 & .00 & .00 & .00 & .00 & .00 & .00 & .00 & .00 & .00 & .00 & .00 & .00 & .00 & .00 & .00 & .00 \\*
 & Few-Shot + CoT & .00 & .00 & .00 & .00 & .00 & .00 & .00 & .00 & .00 & .00 & .00 & .00 & .00 & .00 & .00 & .00 & .00 & .00 & .00 & .00 & .00 & .00 & .00 & .00 \\
\end{longtable}
\endgroup

\section{The prompt}
\label{app:prompt}

Every judge receives the same prompt, assembled from three fragments so the
experimental factors stay orthogonal: a \textsc{base} block that never varies, an
image clause that appears only when an image is attached, and an output block
fixed by whether chain-of-thought is requested. Below is the complete
\emph{zero-shot, explicit} configuration, the one whose image clause states the
guideline rule that the image is context only. Markdown emphasis is rendered
here; the model receives the raw text.

\begin{center}
\footnotesize
\setlength{\fboxsep}{7pt}
\fbox{\begin{minipage}{0.93\linewidth}
\noindent{\small\textbf{\textsf{SYSTEM}}}\par\smallskip
\noindent{\scriptsize\textsf{\textcolor{blue!45!black}{\textsc{base} --- identical in every condition}}}\par\smallskip
You are an expert linguistic annotator working on a figurative language annotation task.

You are given one sentence containing a highlighted phrase. The phrase is wrapped in double asterisks, like \textbf{this}. Your task is to decide how that phrase is being used in the exact sentence given.

You are labeling the phrase in the sentence. You are not labeling anything else.

\smallskip\noindent\textbf{\textsf{Label definitions}}\par\smallskip

There are exactly four labels.

\textbf{Fully Figurative (FF)} --- The phrase is used in a completely figurative sense. Its literal meaning plays no role whatsoever in the sentence. There is zero connection between what the words literally refer to and how they are used here. Example: "After a long illness, the elderly man finally \textbf{kicked the bucket}, surrounded by his loving family." --- means 'died'; kicking and buckets are entirely irrelevant.

\textbf{Weak Figurative (WF)} --- The phrase is used figuratively in the sentence, but there is still a meaningful semantic or conceptual link between the figurative meaning and the literal words. The connection may be abstract, metaphorical, spatial, structural, or a shared term. Example: "The team was \textbf{racing against the clock} to meet the project deadline." --- means working under time pressure; a clock measures time, so the literal words and the figurative meaning are directly linked.

\textbf{Figurative and Literal (FL)} --- The phrase functions on both levels simultaneously in this sentence: the literal meaning is genuinely activated AND the figurative meaning is clearly present. Both readings must be plausible and actually present, not merely theoretically possible. Example: "The detective was \textbf{leaving no stone unturned} in her investigation, following every lead and interviewing every witness." --- figuratively 'doing everything possible', while at a crime scene stones could plausibly be physically turned over.

\textbf{Fully Literal (LL)} --- The phrase is used purely in its word-by-word literal sense. No figurative reading is activated in this sentence. Example: "The blacksmith carefully handled the \textbf{red hot} metal rod, shaping it into a perfect horseshoe." --- the rod is literally red and literally hot; the 'highly popular' sense is not invoked.

\smallskip\noindent\textbf{\textsf{Decision rules --- follow in order}}\par\smallskip

\textbf{Step 1 --- Literal check (depends on the sentence).} Ask: does the phrase play its word-by-word literal meaning in this exact sentence?
\noindent\hspace*{1em}$\bullet$~Yes, and no figurative meaning is present $\rightarrow$ \textbf{LL}\par
\noindent\hspace*{1em}$\bullet$~Yes, and a figurative meaning is also present $\rightarrow$ \textbf{FL}\par
\noindent\hspace*{1em}$\bullet$~No $\rightarrow$ go to Step 2\par

The literal reading must be plausible in the sentence as written, not merely theoretically possible.

\textbf{Step 2 --- Figurative type check (ignores the sentence).} Ask: ignoring the sentence context entirely, does the figurative meaning have ANY semantic or conceptual connection to what the words literally mean?
\noindent\hspace*{1em}$\bullet$~Yes, any connection at all, even abstract $\rightarrow$ \textbf{WF}\par
\noindent\hspace*{1em}$\bullet$~No connection at all $\rightarrow$ \textbf{FF}\par

The connection can be a shared term with a related meaning, a spatial or physical metaphor, a sequence or structural analogy, or a direct conceptual link.

\smallskip\noindent\textbf{\textsf{Additional rules}}\par\smallskip

\noindent\hangindent=1.2em 1. \textbf{Be strict with the exact sentence grammar.} Annotate the sentence exactly as written. Do not add or remove articles, do not substitute near-synonyms, and do not assume a different grammatical form. If the phrase would be literal only with a slight grammar change, it is not literal as written.\par
\noindent\hangindent=1.2em 2. \textbf{Read the full sentence before deciding.} One adjective, subject, or clause elsewhere in the sentence can determine whether the literal reading is plausible. A sentence about an actual beaver named Benny building a dam makes "eager beaver" FL, not FF.\par
\noindent\hangindent=1.2em 3. \textbf{Step 1 depends on context; Step 2 ignores it.} Step 1 is a judgment about this sentence. Step 2 is a judgment about the phrase in general --- the relationship between its literal words and its figurative meaning.\par
\noindent\hangindent=1.2em 4. \textbf{Do not assume famous idioms are figurative.} A well-known idiom can be LL, FL, WF, or FF depending on the sentence. Always run Step 1 on the actual words in front of you.\par
\noindent\hangindent=1.2em 5. \textbf{If you are unsure between two labels}, choose the one that best describes your primary interpretation. If the uncertainty comes from the phrase genuinely supporting two readings at once, that is a signal to choose FL.\par\smallskip\noindent\textcolor{gray}{\rule{\linewidth}{0.4pt}}\par\vspace{2pt}\noindent{\scriptsize\textsf{\textcolor{blue!45!black}{\textsc{image clause} --- present only when an image is attached; this is the \emph{explicit} arm}}}\par\smallskip
\smallskip\noindent\textbf{\textsf{The image}}\par\smallskip

An image is shown alongside the sentence as silent background context. It represents a scene or setting; it is not your annotation target.

Do not use the image to change your interpretation of the phrase. You are labeling the phrase in the sentence, not judging whether the image matches the phrase. A literal-looking image does not make the phrase Fully Literal, and a figurative image does not make it figurative --- you must still read the sentence.

If the image is distracting, annotate as if the image were hidden, then check that your answer still holds.\par\smallskip\noindent\textcolor{gray}{\rule{\linewidth}{0.4pt}}\par\vspace{2pt}\noindent{\scriptsize\textsf{\textcolor{blue!45!black}{\textsc{output} --- the zero-shot and few-shot form; the CoT form also asks for \texttt{thought\_process}}}}\par\smallskip
\smallskip\noindent\textbf{\textsf{Output}}\par\smallskip

Apply the decision rules silently. Do not write out your reasoning.

Respond with a single JSON object and nothing else:

\noindent\texttt{\{"final\_label": "FF" \textbar{} "WF" \textbar{} "FL" \textbar{} "LL"\}}
\end{minipage}}

\vspace{5pt}

\setlength{\fboxsep}{7pt}
\fbox{\begin{minipage}{0.93\linewidth}
\noindent{\small\textbf{\textsf{USER}}}\par\smallskip
\noindent\textit{[the target image, attached as a JPEG data URI, precedes the text]}\par\smallskip
\noindent\texttt{Phrase: 'Hit the sack'}\par
\noindent\texttt{Sentence: 'After a long day, John was ready to **hit the}\par
\noindent\texttt{sack** and get some much-needed rest.'}
\end{minipage}}
\end{center}

Under \emph{few-shot} the four demonstrations are inserted between the system
turn and this user turn, as alternating text-only user and assistant messages.
They are drawn deterministically, and therefore identically across judges, from
twenty items the informed trio labelled unanimously, one per label, and never
include the target item. Under the \emph{silent} arm the image clause is absent
and nothing else changes.



\end{document}